\RequirePackage{fix-cm}
\documentclass[twocolumn]{svjour3}          
\smartqed  
\usepackage{graphicx}
\usepackage{times}
\usepackage{epsfig}
\usepackage{bbm}
\usepackage{amsmath}
\usepackage{amssymb}
\usepackage{subfigure}
\usepackage{color,soul}
\usepackage{url}
\usepackage{natbib}
\usepackage{marvosym}
\begin{document}

\title{LIBSVX: A Supervoxel Library and Benchmark\\for Early Video Processing
}


\author{
Chenliang Xu \and
Jason J. Corso
}


\institute{
C. Xu (\Letter) \and J. J. Corso \at
Electrical Engineering and Computer Science\\
University of Michigan,\\
1301 Beal Avenue \\
Ann Arbor, MI 48109-2122, USA\\
\email{cliangxu@umich.edu}
\and
J. J. Corso\\
\email{jjcorso@eecs.umich.edu}
}

\date{Received: date / Accepted: date}

\maketitle

\begin{abstract}
Supervoxel segmentation has strong potential to be incorporated into early video 
analysis as superpixel segmentation has in image analysis. However, there are 
many plausible supervoxel methods and little understanding as to when and where 
each is most appropriate. Indeed, we are not aware of a single comparative study 
on supervoxel segmentation. To that end, we study seven supervoxel algorithms, 
including both off-line and streaming methods, in the context of what we 
consider to be a good supervoxel: namely, spatiotemporal uniformity, 
object/region boundary detection, region compression and parsimony. For the 
evaluation we propose a comprehensive suite of seven quality metrics to measure 
these desirable supervoxel characteristics. In addition, we evaluate the methods 
in a supervoxel classification task as a proxy for subsequent high-level uses of 
the supervoxels in video analysis. We use six existing benchmark video datasets 
with a variety of content-types and dense human annotations. Our findings have 
led us to conclusive evidence that the hierarchical graph-based (GBH), 
segmentation by weighted aggregation (SWA) and temporal superpixels (TSP) 
methods are the top-performers among the seven methods. They all perform well in 
terms of segmentation accuracy, but vary in regard to the other desiderata: GBH 
captures object boundaries best; SWA has the best potential for region 
compression; and TSP achieves the best 
undersegmentation error.


\end{abstract}

\section{Introduction}
\label{sec:introduction}

Images have many pixels; videos have more. It has thus become standard practice to first preprocess images and videos into more tractable sets by either extraction of salient points \citep{ScMoTPAMI1997} or oversegmentation into superpixels \citep{ReMaICCV2003}. Preprocessing output---salient points or superpixels---is more perceptually meaningful than raw pixels, which are merely a consequence of digital sampling \citep{ReMaICCV2003}. However, the same practice does not entirely exist in video analysis. Although many methods do indeed initially extract salient points or dense trajectories, e.g., \cite{LaIJCV2005}, we are aware of few methods that rely on a supervoxel segmentation, which is the video analog to a superpixel segmentation. In fact, those papers that do preprocess video tend to rely on a per-frame superpixel segmentation, e.g., \cite{LeKiGrICCV2011}, or use a full-video segmentation, e.g., \cite{GrKwHaCVPR2010}.

The basic position of this paper is that supervoxels have great potential in 
advancing video analysis methods, as superpixels have for image analysis. To 
that end, we perform a thorough comparative evaluation of seven supervoxel 
methods: five off-line and two streaming methods. The off-line methods require 
the video to be available in advance and short enough to fit in memory. They 
load the whole video at once and process it afterwards. The five off-line 
methods we choose---segmentation by weighted aggregation (SWA) 
\citep{ShBrBaCVPR2000,ShGaShNATURE2006,CoShDuTMI2008}, graph-based (GB) 
\citep{FeHuIJCV2004}, hierarchical graph-based (GBH) \citep{GrKwHaCVPR2010}, 
mean shift \citep{PaDuCVPR2007}, and Nystr\"{o}m normalized cuts (NCut) 
\citep{FoBeChTPAMI2004,ShMaTPAMI2000,FoBeMaCVPR2001}---broadly sample the 
methodology-space, and are intentionally selected to best analyze methods with 
differing qualities for supervoxel segmentation. For example, both SWA and NCut 
use the normalized cut criterion as the underlying objective function, but SWA 
minimizes it hierarchically whereas NCut does not. Similarly, there are two 
graph-based methods that optimize the same function, but one is subsequently 
hierarchical (GBH). We note that, of the off-line methods, only GBH had been 
proposed intrinsically as a supervoxel method; each other one is either 
sufficiently general to serve as one or has been adapted to serve as one. We 
also note a similar selection of segmentation methods has been used in the (2D) 
image boundary comparative study \citep{ArMaFoTPAMI2011} and nonetheless our 
selections share a good overlap with the ones studied in the recent video segmentation benchmark \citep{GaNaCaICCV2013}.

In contrast, streaming methods require only constant memory (depends on the 
streaming window range) to execute the algorithm which makes them feasible for 
surveillance or to run over a long video on a less powerful machine. The two 
streaming methods we choose---streaming hierarchical video segmentation 
(streamGBH) \citep{XuXiCoECCV2012} and temporal superpixels (TSP) 
\citep{ChWeIICVPR2013} employ different strategies to treat video data. The 
streamGBH approximates a full video segmentation by both hierarchical and 
temporal Markov assumptions. Each time it segments video frames within a 
streaming window, and the length of the streaming window can be as short as one 
frame or as long as the full video, which equates it to full-video GBH 
segmentation. TSP represents a set of methods 
\citep{ChWeIICVPR2013,VaRoBoICCV2013,ReJaRoICCV2013} that computes the superpixel 
segmentation on the first frame and then extends the superpixels to subsequent 
frames (one by one) in a video.  The TSP method \citep{ChWeIICVPR2013} uses a 
Gaussian Process for the streaming segmentation.

Our paper pits the five off-line and two streaming methods in an evaluation on a 
suite of  metrics designed to assess the methods on basic supervoxel desiderata 
(Sect.~\ref{sec:good}), such as following object boundaries and spatiotemporal 
coherence. The specific metrics we use are 3D undersegmentation error, 3D 
segmentation accuracy, boundary recall distance and label consistency. They 
evaluate the supervoxel segmentations against human annotations. We also use a 
set of human-independent metrics: explained variation, mean size variation and 
temporal extent of supervoxels, which directly explore the properties of each 
method. Finally, we compare the supervoxel methods in a particular 
application---supervoxel classification---that evaluates methods in a 
recognition task, which we consider to be a proxy to various high-level video 
analysis tasks in which supervoxels could be used.  We use six complementary 
video datasets to facilitate the study:  BuffaloXiph 
\citep{ChCoWNYIPW2010}, SegTrack v2 \citep{TsFlReBMVC2010,LiKiHuICCV2013}, 
CamVid \citep{BrShFaECCV2008}, BVDS \citep{SuBrMaCVPR2011,GaNaCaICCV2013}, 
\cite{LiFrAdCVPR2008} and Middlebury Flow \citep{BaScLeIJCV2011}. They span from 
few videos to one hundred videos, and from sparse annotations to dense 
frame-by-frame annotations.

A preliminary version of our work appears in \cite{XuCoCVPR2012}. Since its 
initial release, the LIBSVX benchmark has been widely used in supervoxel method 
evaluation by the community, including but not limited to 
\cite{XuXiCoECCV2012,XuWhCoICCV2013,ChWeIICVPR2013,PaSaCVPR2013,ReJaRoICCV2013,VaRoBoICCV2013,LeChAISTAT2014,SoAlotPRL2014,TrHwBeWACV2014}. 
In this paper, we complement the library with the two streaming methods and a set of new benchmark metrics on new video datasets. In addition, we add a new experiment of supervoxel classification to evaluate methods in terms of a middle-level video representation towards high-level video analysis. We also note that a recent video segmentation evaluation is proposed in \cite{GaNaCaICCV2013}. We distinguish our work from them by evaluating directly on supervoxel segmentation, the oversegmentation of a video, and using various datasets including densely labeled human annotations with a set of novel benchmark metrics. It is our position that evaluations of both over-segmentation and segmentation in video are necessary to establish a thorough understanding of the problem-space within the computer vision community.

Our evaluation yields conclusive evidence that GBH, SWA and TSP are the top-performers 
among the seven methods. They all perform well in terms of segmentation 
accuracy, but vary in regard to the other desiderata:
GBH captures object boundaries best; SWA 
has the best potential for region compression; and TSP follows object parts and 
achieves the best undersegmentation error.  Although GBH and SWA, the two 
offline methods, are quite distinct in formulation and may perform differently 
under other assumptions, we find a common feature among the two methods (and one 
that separates them from the others) is the manner in which coarse level 
features are incorporated into the hierarchical computation. TSP is the only 
streaming method among the three and generates supervoxels with the best 
spatiotemporal uniformity. Finally, the supervoxel classification experiment 
further supports our findings and shows a strong correlation to our benchmark 
evaluation.

The complete supervoxel library, benchmarking code, classification code and documentation are available for download at \url{http://www.supervoxels.com}. Various supervoxel results on major datasets in the community (including the existing six datasets \cite{ChCoWNYIPW2010,TsFlReBMVC2010,LiKiHuICCV2013,BrShFaECCV2008,SuBrMaCVPR2011,GaNaCaICCV2013,LiFrAdCVPR2008,BaScLeIJCV2011}) are also available at this location to allow for easy adoption of the supervoxel results by the community.

The rest of the paper is organized as follows. We present a theoretical background in Sect.~\ref{sec:background} and a brief description of the methods in Sect.~\ref{sec:methods}. We introduce the datasets and processing setup in Sect.~\ref{sec:datasets}. We thoroughly discuss comparative performance in terms of benchmark in Sect.~\ref{sec:benchmark} and supervoxel classification in Sect.~\ref{sec:classification}. Finally, we conclude the paper in Sect.~\ref{sec:conclusion}.

\section{Background}
\label{sec:background}

\subsection{Superpixels}

The term \textit{superpixel} was coined by \cite{ReMaICCV2003} in their work on learning a binary classifier that can segment natural images. The main rationale behind superpixel oversegmentation is twofold: (1) pixels are not natural elements but merely a consequence of the discrete sampling of the digital images and (2) the number of pixels is very high making optimization over sophisticated models intractable. \cite{ReMaICCV2003} use the normalized cut algorithm \citep{ShMaTPAMI2000} for extracting the superpixels, with contour and texture cues incorporated. Subsequently, many superpixel methods have been proposed \cite{LeStKuTPAMI2009,VeBoMeECCV2010,MoPrWaCVPR2008,LiTuRaCVPR2011,ZeWaWaICCV2011} or adopted as such \citep{FeHuIJCV2004,ViSoTPAMI1991,CoMeTPAMI2002} and used for a variety of applications: e.g., human pose estimation \citep{MoReEfCVPR2004}, semantic pixel labeling \citep{HeZeRaECCV2006,TiLaIJCV2010}, 3D reconstruction from a single image \citep{HoEfHeTOG2005} and multiple-hypothesis video segmentation \citep{VaAvPfECCV2010} to name a few. Few superpixel methods have been developed to perform well on video frames, such as \cite{DrMaWMVC2009} who base the method on minimum cost paths but do not incorporate any temporal information.


\subsection{What makes a good supervoxel method?}
\label{sec:good}

First, we define a \textit{supervoxel}---the video analog to a superpixel. Concretely, given a 3D lattice $\Lambda^3$ (the voxels in the video), a supervoxel $v$ is a subset of the lattice $v \subset \Lambda^3$ such that the union of all supervoxels comprises the lattice and they are pairwise disjoint: $\bigcup_i v_i = \Lambda^3 \wedge v_i \bigcap v_j = \varnothing \; \forall i,j \text{ pairs}$. Obviously, various image/video features may be computed on the supervoxels, such as color histograms and textons. In this initial definition, there is no mention of certain desiderata that one may expect, such as locality, coherence, and compactness. Rather than include them in mathematical terms, we next list terms of this sort as desirable characteristics of a \textit{good} supervoxel method.  

We define a good supervoxel method based jointly on criteria for good supervoxels, which follow closely from the criteria for good segments \citep{ReMaICCV2003}, and the actual cost of generating them (videos have an order of magnitude more pixels over which to compute). Later, in our experimental evaluation, we propose a suite of benchmark metrics designed to evaluate these criteria (Section \ref{sec:benchmark}).

\textbf{Spatiotemporal Uniformity.} The basic property of spatiotemporal uniformity, or \textit{conservatism} \citep{MoPrWaCVPR2008}, encourages compact and uniformly shaped supervoxels in space-time \citep{LeStKuTPAMI2009}. This property embodies many of the basic Gestalt principles---proximity, continuation, closure, and symmetry---and helps simplify computation in later stages \citep{ReMaICCV2003}. Furthermore, \cite{VeBoMeECCV2010} show that for the case of superpixels, compact segments perform better than those varying in size on the higher level task of salient object segmentation. For temporal uniformity (called coherence in \cite{GrKwHaCVPR2010}), we expect a mid-range compactness to be most appropriate for supervoxels (bigger than, say, five frames and less than the whole video).

\textbf{Spatiotemporal Boundaries and Preservation.} The supervoxel boundaries should align with object/region boundaries when they are present and the supervoxel boundaries should be stable when they are not present; i.e., the set of supervoxel boundaries is a superset of object/region boundaries. Similarly, every supervoxel should overlap with only one object \citep{LiTuRaCVPR2011}. Furthermore, the supervoxel boundaries should encourage a high-degree of \textit{explained variation} \citep{MoPrWaCVPR2008} in the resulting oversegmentation. If we consider the oversegmentation by supervoxels as a compression method in which each supervoxel region is represented by the mean color, we expect the distance between the compressed and original video to have been minimized.

\textbf{Computation.} The computation cost of the supervoxel method should reduce the overall computation time required for the entire application in which the supervoxels are being used.

\textbf{Performance.} The oversegmentation into supervoxels should not reduce the achievable performance of the application. Our evaluation will not directly evaluate this characteristic (because we study the more basic ones above).

\textbf{Parsimony.} The above properties should be maintained with as few supervoxels as possible \citep{LiTuRaCVPR2011}.

\section{Methods}
\label{sec:methods}

We study seven supervoxel methods---mean shift \citep{PaDuCVPR2007}, graph-based (GB) \citep{FeHuIJCV2004}, hierarchical graph-based (GBH) \citep{GrKwHaCVPR2010}, streaming hierarchical graph-based (streamGBH) \citep{XuXiCoECCV2012}, Nystr\"{o}m normalized cut (NCut) \citep{ShMaTPAMI2000,FoBeMaCVPR2001,FoBeChTPAMI2004}, segmentation by weighted aggregation (SWA) \citep{ShBrBaCVPR2000,ShGaShNATURE2006,CoShDuTMI2008} and temporal superpixels \citep{ChWeIICVPR2013}---that broadly sample the methodology-space among statistical and graph partitioning methods \citep{ArMaFoTPAMI2011}. We have selected these seven due to their respective traits and their inter-relationships: for example, Nystr\"{o}m and SWA both optimize the same normalized cut criterion, and streamGBH extends GBH to handle arbitrarily long videos and still keeps the hierarchy property.

We describe the methods in some more detail below. We note that \textit{many} other methods have been proposed in the computer vision literature for video segmentation, e.g., \cite{ViSoTPAMI1991,GrGoMaTPAMI2004,BrToICCV2009,LiDoYaCVPR2008,VaAvPfECCV2010,VeBoMeECCV2010,KhShCVPR2001,MeDeUMD2002,BuBaCiCVPR2011,GaCiScACCV2012}, but we do not cover them in any detail in this study. We also do not cover strictly temporal segmentation, e.g. \cite{PaSePR1997}.

%
%
%
%
%

\subsection{Mean Shift}

Mean shift is a mode-seeking method, first proposed by \cite{FuHoTIT1975}. \cite{CoMeTPAMI2002} and \cite{WaThXuECCV2004} adapt the kernel to the local structure of the feature points, which is more computationally expensive but improves segmentation results. Original hierarchical mean shift in video \citep{DeMeSMVPW2002,PaECCV2008} improves the efficiency of (isotropic) mean-shift methods by using a streaming approach. The mean shift algorithm used in our paper is presented by \cite{PaDuCVPR2007}, who introduce Morse theory to interpret mean shift as a topological decomposition of the feature space into density modes. A hierarchical segmentation is created by using topological persistence. Their algorithm is more efficient than previous works especially on videos and large images. We use the author-provided implementation\footnote{\url{http://people.csail.mit.edu/sparis/}} to generate a supervoxel hierarchy and then stratify the pairwise merging into a fixed-level of hierarchy.

\subsection{Graph-Based (GB)}

\cite{FeHuIJCV2004} propose a graph-based algorithm for image segmentation; it is arguably the most popular superpixel segmentation method. Their algorithm runs in time nearly linear in the number of image pixels, which makes it suitable for extension to spatiotemporal segmentation. Initially, each pixel, as a node, is placed in its own region $R$, connected with 8 neighbors. Edge weights measure the dissimilarity between nodes (e.g. color differences). They define the internal difference of a region, $Int(R)$, as the largest edge weight in the minimum spanning tree of $R$. Traversing the edges in a non-decreasing weight order, the regions $R_i$ and $R_j$ incident to the edge are merged if the current edge weight is less than the relaxed minimum internal difference of the two regions:
\begin{align}
\min(Int(R_i)+\tau(R_i),Int(R_j)+\tau(R_j))
\enspace,
\end{align}
where $\tau(R) = k/|R|$ is used to trigger the algorithm and gradually makes it converge. $k$ is a scale parameter that reflects the preferred region size. The algorithm also has an option to enforce a minimum region size by iteratively merging low-cost edges until all regions contain the minimum size of pixels. We have adapted the algorithm for video segmentation by building a 3D lattice over the spatiotemporal volume, in which voxels are nodes connected with 26 neighbors in the lattice (9 to the previous and the next frames, 8 to the current frame). One challenge in using this algorithm is the selection of an appropriate $k$ for a given video, which the hierarchical extension (GBH, next) overcomes. We use a set of $k$ as well as various minimum region sizes to generate the segmentation output for our experiment.

\subsection{Hierarchical Graph-Based (GBH)}

The hierarchical graph-based video segmentation algorithm is proposed by \cite{GrKwHaCVPR2010}. Their algorithm builds on an oversegmentation of the above spatiotemporal graph-based segmentation. It then iteratively constructs a region graph over the obtained segmentation, and forms a bottom-up hierarchical tree structure of the region (segmentation) graphs. Regions are described by local Lab histograms. At each step of the hierarchy, the edge weights are set to be the $\chi^2$ distance between the Lab histograms of the connected two regions. They apply the same technique as above, \cite{FeHuIJCV2004}, to merge regions. Each time they scale the minimum region size as well as $k$ by a constant factor $s$. Their algorithm not only preserves the important region borders generated by the oversegmentation, but also allows a selection of the desired segmentation hierarchy level $h$, which is much better than directly manipulating $k$ to control region size. We set a large $h$ to output segmentations with various numbers of supervoxels.

\subsection{Graph-Based Streaming Hierarchical (streamGBH)}
\label{subsec:streamGBH}

Graph-based streaming hierarchical video segmentation is proposed in our earlier 
work \citep{XuXiCoECCV2012} to extend GBH \citep{GrKwHaCVPR2010} to handle 
arbitrarily long videos in a streaming fashion and still maintain the 
segmentation hierarchy. 
The algorithm approximates the full video GBH 
segmentations by both a hierarchical and a temporal Markov assumption, allowing 
a small number of frames to be loaded into a memory at any given time. Therefore the algorithm 
runs in a streaming fashion. In our comparison experiments, we set a fixed 
streaming window size (10 frames) for all subsequences and, again, a large $h$ as 
in GBH to output segmentations with various numbers of supervoxels.

\subsection{Nystr\"{o}m Normalized Cut (NCut)}

Nystr\"{o}m Normalized Cuts \citep{ShMaTPAMI2000} as a graph partitioning criterion has been widely used in image segmentation. A multiple eigenvector version of normalized cuts is presented in \cite{FoBeChTPAMI2004}. Given a pairwise affinity matrix $W$, they compute the  eigenvectors $V$ and eigenvalues $\Gamma$ of the system
\begin{align}
(D^{-1/2}WD^{-1/2})V = V\Gamma
\enspace,
\end{align}
where $D$ is a diagonal matrix with entries $D_{ii} = \sum_j W_{ij}$. Each voxel is embedded in a low-dimensional Euclidean space according to the largest several eigenvectors. The k-means algorithm is then be used to do the final partitioning. To make it feasible to apply to the spatiotemporal video volume, \cite{FoBeMaCVPR2001} use the Nystr\"{o}m approximation to solve the above eigenproblem. Their paper demonstrates segmentation on relatively low-resolution, short videos (e.g., $120 \times 120 \times 5$) and randomly samples points from the first, middle, and last frames.

However, in our experiments, NCut is not scalable as the number of supervoxels and the length of video increases. Sampling too many points makes the Nystr\"{o}m method require too much memory, while sampling too few gives unstable and low performance. Meanwhile, the k-means clustering algorithm is sufficient for a video segmentation with few clusters, but a more efficient clustering method is expected regarding the number of supervoxels. Therefore, we run NCut for a subset of our experiments with lower solution and we set 200 sample points. We run k-means on 20\% of the total voxels and k-nearest neighbor search to assign supervoxel labels for all voxels.

\subsection{Segmentation by Weighted Aggregation (SWA)}

SWA is an alternative approach to optimizing the normalized cut criterion \citep{ShBrBaCVPR2000,ShGaShNATURE2006,CoShDuTMI2008} that computes a hierarchy of sequentially coarser segmentations. The method uses an algebraic multigrid solver to compute the hierarchy efficiently. It recursively coarsens the initial graph by selecting a subset of nodes such that each node on the fine level is \textit{strongly coupled} to one on the coarse level. The algorithm is nearly linear in the number of input voxels, and produces a hierarchy of segmentations, which motivates its extension to a supervoxel method. The SWA implementation is based on our earlier 3D-SWA work in the medical imaging domain \citep{CoShDuTMI2008}.  

\subsection{Temporal Superpixels (TSP)}

The temporal superpixels method computes the superpixel segmentation on the first frame and then extends the existing superpixels to subsequent frames in a video. Therefore, this set of methods \cite{ChWeIICVPR2013,VaRoBoICCV2013,ReJaRoICCV2013}, by their nature, are computing supervoxels in a streaming fashion, which is similar to streamGBH with a streaming window of one frame. We choose \cite{ChWeIICVPR2013} as the representative method for evaluation. The algorithm first extends the SLIC \citep{AcShSmTPAMI2012} superpixel algorithm to form a generative model for constructing superpixels. Each pixel is modeled using five dimensional feature vector: three channel color and the 2D location in image. Superpixels are inferred by clustering with a mixture model on individual features as a Gaussian with known variance. After generating superpixels for the first frame, the algorithm applies a Gaussian Process with a bilateral kernel to model the motion between frames. We use the implementation\footnote{\url{http://people.csail.mit.edu/jchang7/code.php}} provided by the authors with the default parameters to run the algorithm in evaluation.

\section{Datasets}
\label{sec:datasets}

We make use of six video datasets for our experimental purposes, with varying 
characteristics. The datasets have human-annotator drawn groundtruth labels at a 
frame-by-frame basis (four out of six) or at densely sampled frames in the video 
(two out of six). The sizes of the selected datasets vary from a few videos to one 
hundred videos. The set of datasets we choose are BuffaloXiph 
\citep{ChCoWNYIPW2010}, SegTrack v2 \citep{LiKiHuICCV2013,TsFlReBMVC2010}, BVDS 
\citep{SuBrMaCVPR2011,GaNaCaICCV2013}, CamVid \citep{BrShFaECCV2008}, 
\cite{LiFrAdCVPR2008} and Middlebury Flow \citep{BaScLeIJCV2011}. The datasets 
are originally built solving different video challenges: BuffaloXiph is 
gathered for pixel label propagation in videos; SegTrack is built for object 
tracking; BVDS has contributed to occlusion boundary detection; CamVid is taken 
in driving cars for road scene understanding; and \cite{LiFrAdCVPR2008} and 
Middlebury Flow \citep{BaScLeIJCV2011} are used for optical flow estimation.  
Rather than evaluating supervoxel methods on a single dataset, we conduct the 
evaluation on all six datasets (with only label consistency metric on 
\cite{LiFrAdCVPR2008} and Middlebury Flow \citep{BaScLeIJCV2011}), as the 
datasets are complementary and we believe supervoxels have potential to be a 
first processing step towards various video applications and problems. We 
briefly describe the six datasets used in our experiments.

\noindent \textbf{BuffaloXiph} from \cite{ChCoWNYIPW2010} is a subset of the well-known \url{xiph.org} videos that have been supplemented with a 24-class semantic pixel labeling set (the same classes from the MSRC object segmentation dataset \cite{ShWiRoIJCV2009}). The eight videos in this set are densely labeled with semantic pixels that leads to a total of 638 labeled frames, with a minimum of 69 frames-per-video (fpv) and a maximum of 86 fpv. The dataset is originally used for pixel label propagation \citep{ChCoWNYIPW2010} and videos in the dataset are stratified according to camera motion, object motion, the presence of articulated objects, the complexity of occlusion between objects and the difficulty of label propagation. Distinct regions with the same semantic class label are not separated in this dataset. 

\noindent \textbf{SegTrack v2} from \cite{LiKiHuICCV2013} is an updated version of the \textit{SegTrack} dataset \citep{TsFlReBMVC2010} and provides frame-by-frame pixel-level  foreground objects labeling rather than the semantic class labeling as in BuffaloXiph. It contains a total of 14 video sequences with 24 objects over 947 annotated frames. The videos in the dataset are stratified according to different segmentation challenges, such as motion blur, appearance change, complex deformation, occlusion, slow motion and interacting objects.

\noindent \textbf{BVDS} is initially introduced in \cite{SuBrMaCVPR2011} for 
occlusion boundary detection and then used for evaluating video segmentation 
algorithms by \cite{GaNaCaICCV2013}. It consists of 100 HD quality videos with a 
maximum of 121 fpv and videos in the dataset are stratified according to 
occlusion, object categories and sizes, and different kinds of camera motion: 
translational, scaling and perspective motion. Each video is labeled with 
multiple human annotations by a sampling rate of 20 frames. We use all 100 
videos in the evaluation ignoring the training/testing split (because BVDS is 
used only in the unsupervised parts of our evaluation). 

Furthermore, the dataset has three different groupings for videos with moving 
objects, non-rigid motion, and considerable camera motion. Our experimental 
results show that all methods preserve the same performance order over these 
three video groupings, except TSP has better temporal extent than GB when only using videos with considerable camera motion. We show this additional result in the supplement.

\noindent \textbf{CamVid} from \cite{BrShFaECCV2008} provides five long video sequences recorded at daytime and dusk from a car driving through Cambridge, England. The videos are composed by over ten minutes high quality 30Hz footage and are labeled with 11 semantic object class labels at 1Hz and in part 15Hz that leads to a total of 701 densely labeled frames. It also provides the training/test split, with two daytime and one dusk sequence for training and one daytime and one dusk sequence for testing. Therefore, this dataset in addition allows us to evaluate methods in terms of supervoxel semantic label classification. We use all videos, in total 17898 frames\footnote{We manually exclude the corrupted frames, and organize the dataset into short clips with roughly 100 frames-per-clip. The organized short clips can be downloaded from our website.}, in the evaluation in Sect.~\ref{sec:benchmark}, and follow the training/test split in Sect.~\ref{sec:classification}.

The remaining two datasets, \textbf{\cite{LiFrAdCVPR2008}} and 
\textbf{Middlebury Flow} \citep{BaScLeIJCV2011} are used for evaluating label 
consistency in Sect.~\ref{subsec:LC}. They are densely annotated with 
groundtruth flows. \cite{LiFrAdCVPR2008} contains five videos with a minimum of 
14 fpv and a maximum of 76 fpv. Middlebury Flow contains eight videos, but 
groundtruth for only two frames (one optical flow estimate) is available. We 
treat it as a special case where algorithms only process two frames.

\subsection{Processing}
\label{subsec:processing}

To adapt all seven supervoxel methods to run through all videos in the datasets within reasonable time and memory consumption, we use BuffaloXiph, SegTrack v2 and Middlebury Flow at the original resolution; \cite{LiFrAdCVPR2008} at half the original resolution; BVDS and CamVid, the two large datasets, at a quarter of the original HD resolution. One exception is NCut which runs at a fixed resolution of $240 \times 160$ on BuffaloXiph and SegTrack v2 datasets (the results are scaled up for comparison) and is not included in the experiments with BVDS and CamVid datasets due to its high computational demands. The comparison of NCut and other methods at the same downscaled resolution on BuffaloXiph and SegTrack are shown in our conference version of the paper \citep{XuCoCVPR2012}, where the relative performance is similar to here.

We compare the seven methods as fairly as possible. However, each method has its 
own tunable parameters; we have tuned these parameters strictly to achieve a 
certain desired number of supervoxels per video (or per frame, depending on the 
experiment); parameters are tuned per method per dataset. For hierarchical 
methods, such as GBH, streamGBH, SWA, a single run over a video can generate 
fine-to-coarse multiple levels of supervoxels. For Mean Shift, we tune the 
persistence threshold to get multiple stratified segmentations. For NCut, we 
vary the final step K-means clustering to get a set of supervoxels varying from 
100 to 500 on BuffaloXiph and SegTrack v2. We use the suggested parameters by 
the authors for the two other methods (Mean Shift and TSP) and we provide all 
parameters to reproduce our experiments.

After we have generated a range of supervoxels for each video in a dataset, we use linear interpolation to estimate each methods' metric outputs for each video densely. The performance over a dataset at a certain number of supervoxels is drawn by averaging the interpolated values from all videos at the same number of supervoxels. This strategy can better align videos in a dataset and therefore avoids outliers with too many or too few supervoxels by simply taking averaged number of supervoxels over a dataset, especially when the videos are diverse in a dataset.

\section{Benchmark Evaluation}
\label{sec:benchmark}

Rather than evaluating the supervoxel methods on a particular application, as 
\cite{HaCIARP2008} does for superpixels and image segmentation, in this section 
we directly consider all of the base traits described in Sect.~\ref{sec:good} at 
a fundamental level. We believe these basic evaluations have a great potential 
to improve our understanding of when a certain supervoxel method will perform well.  
Nonetheless, we further evaluate the performances of the supervoxel 
classification on the CamVid dataset in Sect.~\ref{sec:classification}.

We note that some quantitative superpixel evaluation metrics have been recently 
used in 
\cite{MoPrWaCVPR2008,LeStKuTPAMI2009,VeBoMeECCV2010,LiTuRaCVPR2011,ZeWaWaICCV2011}. 
We select those most appropriate to validate our desiderata from Section 
\ref{sec:good}. One way to conduct the experiments is by evaluating the 
frame-based measures and take the average over all the frames in the video. 
However, if we directly apply these methods to the supervoxel segmentation, the 
temporal coherence property can not be captured. Even a  method without any 
temporal information can achieve a good performance in those 2D metrics, which 
have driven us to extend the above frame-based measures to the volumetric 
video-based measures when appropriate. 

In the rest of this section, we first introduce a pair of volumetric video-based 3D metrics that score a supervoxel segmentation based on a given human annotation and they are 3D undersegmentation error (Sect.~\ref{subsec:UE3D}) and 3D segmentation accuracy (Sect.~\ref{subsec:SA3D}). We also evaluate the boundary recall distance of the supervoxel segmentation to the human drawn boundaries (Sect.~\ref{subsec:BRD}), as well as measure the label consistency in terms of annotated groundtruth flows in a video (Sect.~\ref{subsec:LC}). Then we evaluate some basic properties of supervoxel segmentation that do not require human annotation, namely explained variation, mean size variation and temporal extent of supervoxels, in Sect.~\ref{subsec:independent}. We also report the computational cost of each supervoxel method (Sect.~\ref{subsec:cost}). We give visual comparison of the supervoxel segmentations against the groundtruth annotation in Fig.~\ref{fig:montage}. Finally, we discuss our findings in Sect.~\ref{subsec:discussion}.

\subsection{3D Undersegmentation Error (UE3D)}
\label{subsec:UE3D}

Undersegmentation error in image segmentation was proposed in 
\cite{LeStKuTPAMI2009}. It measures the fraction of pixels that exceed the 
boundary of the groundtruth segment when overlapping the superpixels on it. We 
extend this concept to a spatiotemporal video volume to measure the space-time 
\textit{leakage} of supervoxels when overlapping groundtruth segments. Given a 
video segmented into supervoxels $\mathbf{s} = \{s_1, s_2, \dots, s_n\}$ and a 
set of annotated groundtruth segments $\mathbf{g} = \{g_1, g_2, \dots, g_m\}$ in 
video, we define the following UE3D as the average fraction of the voxels that 
exceed the 3D volume of groundtruth segments:
\begin{align}
\text{UE3D}(\mathbf{s}, \mathbf{g}) = \frac{1}{m}
\sum_{i=1}^m \frac{\sum_{j=1}^{n}  \text{Vol}(s_j | s_j \cap g_i \neq \emptyset) 
- \text{Vol}(g_i)}{\text{Vol}(g_i)}
\enspace,
\label{eq:UE3D}
\end{align}
where $\text{Vol}(\cdot)$ denotes the amount of voxels that are contained in the 
3D volume of a segment. Equation \ref{eq:UE3D} takes the average score from all 
groundtruth segments $\mathbf{g}$. We note that the score from a single 
groundtruth segment $g_i$ is not bounded. The metric imposes a greater penalty 
when supervoxels leak on smaller groundtruth segments. For example, if a video 
has a very small object, it will be equally weighted with a large object (e.g.  
background).  Missing a pixel in the small object has a greater penalty than 
missing a background pixel. We also note that it is possible to set different 
weights for groundtruth segment classes when evaluating against a dataset with 
pixel semantic labels (e.g. BuffaloXiph). For a dataset with multiple human 
annotations (e.g. BVDS), we simply take the average score, which equally weights 
different human perceptions. 

\subsection{3D Segmentation Accuracy (SA3D)}
\label{subsec:SA3D}

\begin{figure*}
\begin{center}
\includegraphics[width=\linewidth]{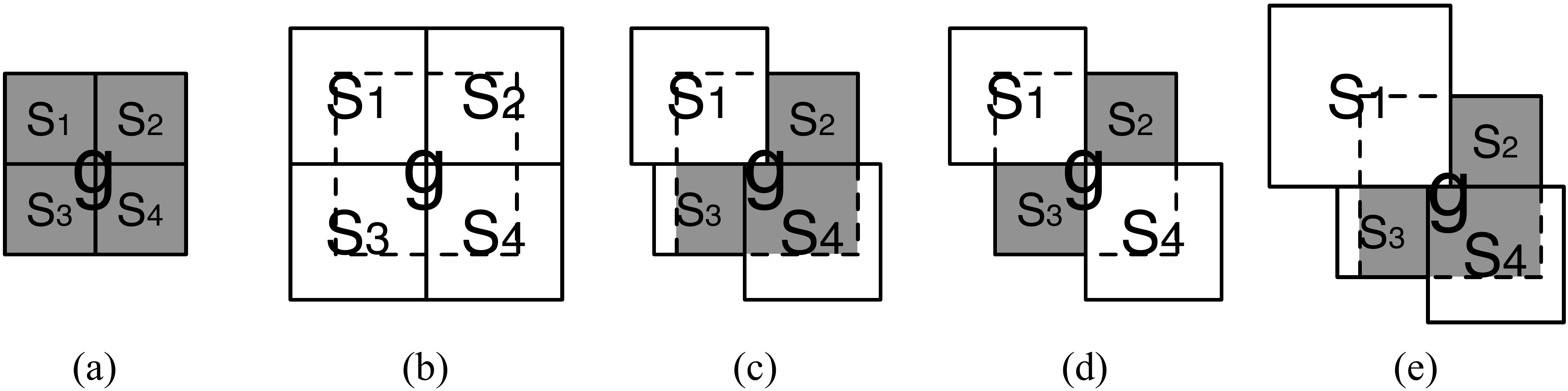}
\caption{A toy example of a single groundtruth segment $g$ with five different 
supervoxel segmentations. We show the example in 2D for simple illustration. We 
draw the groundtruth segment $g$ as a 2x2 dashed square shape. All supervoxel segments are shown in solid square shapes and are defined in three different sizes: 1x1 (e.g. $s_1$ in (a)), 1.5x1.5 (e.g. $s_1$ in (b)), and 2x2 (e.g. $s_1$ in (e)). Segment $s_3$ in (c) and (e) is offset by 1/4. The gray areas are counted toward SA3D.  The scores of UE3D, SA3D and BRD for each cases are shown in Tab.~\ref{tab:BoxScore}.}
\label{fig:Box}
\end{center}
\end{figure*}

Segmentation accuracy measures the average fraction of groundtruth segments that 
is correctly covered by the supervoxels: each supervoxel belongs to only one 
groundtruth segment (object) as a desired property from Sect.~\ref{sec:good}. We 
define the volumetric SA3D as
\begin{align}
&\text{SA3D}(\mathbf{s}, \mathbf{g}) = \nonumber \\ &\frac{1}{m} \sum_{i=1}^m
\frac{ \sum_{j=1}^n \text{Vol}(s_j \cap g_i) \mathbbm{1}[ \text{Vol}(s_j \cap 
g_i) \geq \text{Vol}(s_j \cap \bar{g_i})] }{\text{Vol}(g_i)}
\enspace,
\end{align}
where $\bar{g_i} = \mathbf{g} \setminus \{g_i\}$ and the indicator function 
decides when there is an association of supervoxels between segment $s_j$ and 
groundtruth segment $g_i$. Similar to UE3D, SA3D also takes the average score 
from all groundtruth segments $\mathbf{g}$. However, the score from a single 
groundtruth segment $g_i$ is bounded in $[0,1]$, where the extreme situations 
$1$ and $0$ are respectively define when $g_i$ is perfectly partitioned by a 
set of supervoxels (e.g. Fig.~\ref{fig:Box}(a)), and $g_i$ is completely 
missed (e.g. Fig.~\ref{fig:Box}(b)). 

We note that UE3D and SA3D are complementary to evaluate an algorithm, as UE3D 
measures the leakage of all supervoxels touching a groundtruth segment and SA3D 
measures the fraction of the groundtruth segment that is correctly segmented.  
To further elucidate the differences between UE3D and SA3D, we show a toy 
example in Fig.~\ref{fig:Box} with scores shown in Tab.~\ref{tab:BoxScore}, 
where (c) and (d) have the same UE3D score but different SA3D scores, and (c) 
and (e) have the same SA3D score but different UE3D scores. (c) has the best 
scores for both UE3D and SA3D among all imperfect segmentation cases (b)-(e). 
Both the metrics are evaluated in space-time, such that they penalize 
supervoxels that break not only spatial boundaries but also temporal boundaries 
of the groundtruth segments---a good superpixel method can achieve high 
performance. However, it typically does so with a large number of supervoxels (the temporal extent is only one frame in this case) for the per-video basis. Therefore datasets with dense human annotations, such as BuffaloXiph and SegTrack v2, are more precise in terms of the 3D volumetric measures. 

\subsection{Boundary Recall Distance (BRD)}
\label{subsec:BRD}

So far we have introduced a pair of 3D metrics defined by the set of groundtruth 
segments.  They intrinsically use the groundtruth boundaries for locating the 
volume of the segments. We now directly evaluate the boundary recall distance, 
which measures how well the groundtruth boundaries are successfully retrieved by 
the supervoxel boundaries. We use BRD proposed in \cite{ChWeIICVPR2013} to 
calculate the average distance from points on groundtruth boundaries to the 
nearest ones on supervoxel boundaries frame-by-frame in a video. It does not 
require a fixed amount of dilation for boundary matching as in typical boundary 
recall measures to offset small localization errors. The specific metric is 
defined as follows:
\begin{align}
\text{BRD}(\mathbf{s}, \mathbf{g}) = \frac{1}{\sum_t 
|\mathcal{B}(\mathbf{g}^t)|} \sum_{t=1}^T \sum_{i \in \mathcal{B}(\mathbf{g}^t)} 
\min_{j \in \mathcal{B}(\mathbf{s}^t)} d(i,j)
\enspace,
\label{eq:BRD}
\end{align}
where $\mathcal{B}(\cdot)$ returns the 2D boundaries of segments in a frame, 
$d(\cdot, \cdot)$ is the Euclidean distance between the two arguments, $|\cdot|$ 
denotes the amout of pixels contained by the argument at a frame, $t$ indexes 
frames in a video (e.g. $\mathbf{g}^t$ denotes the set of all groundtruth 
segments on frame $t$), and $i$ and $j$ denote points on boundaries. 

\begin{table}
\begin{center}
\includegraphics[width=\linewidth]{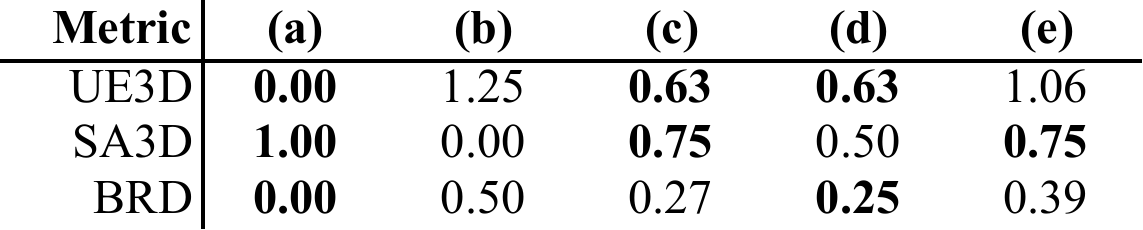}
\caption{The scores of UE3D, SA3D and BRD for the toy example in 
Fig.~\ref{fig:Box}. The larger the better for SA3D, and the small the better for 
UE3D and BRD. The top two scores are bolded for each metric. BRD is calculated strictly for vertical boundary matching only and horizontal boundary matching only in this toy example, which is slightly different than Eq.~\ref{eq:BRD}.}
\label{tab:BoxScore}
\end{center}
\end{table}

We also compute BRD for all cases in Fig.~\ref{fig:Box} and show the scores in 
Tab.~\ref{tab:BoxScore}. We note that BRD captures different aspects of an 
algorithm than UE3D and SA3D. For example, among the imperfect segmentation 
cases (b)-(e) (which are typical situations), (c) has the best scores in terms 
of UE3D and SA3D, but worse in BRD than (d) which is poor in SA3D. Therefore, 
there is no single segmentation that has the best scores for all three metrics 
(except the perfect partition in (a)) in this toy example.

\subsection{Label Consistency (LC)}
\label{subsec:LC}

LC is also proposed in \cite{ChWeIICVPR2013}, which provides a possible way to measure how well supervoxels track the parts of objects given annotated groundtruth flows. Define $\mathcal{F} = \{F^{t-1 \to t} | t = 2, \dots, T\}$ as the vectorized groundtruth forward flow field in a video and $F^{t-1 \to t}(s_i)$ projects pixels contained in $s_i$ at frame $t-1$ to pixels at frame $t$ by the flow (subjected to the image boundary). The metric is defined as follows: 
\begin{align}
\text{LC}(\mathbf{s}, \mathcal{F}) = \frac{\sum_{t=2}^T \sum_{i=1}^n |s_i^t \cap F^{t-1 \to t}(s_i)|}
{\sum_{t=2}^T \sum_{i=1}^n |F^{t-1 \to t}(s_i)|}
\enspace,
\label{eq:LC}
\end{align}
where $s_i^t$ denotes the slice of supervoxel $s_i$ at frame $t$, and the numerator measures the agreement of supervoxel labels and the projected labels by flow. We evaluate this metric on \cite{LiFrAdCVPR2008} and Middlebury Flow where the groundtruth flow annotation is available.

\subsection{Human-Independent Metrics}
\label{subsec:independent}

The following are human-independent metrics; in other words, they are not susceptible to variation in annotator perception that would result in differences in the human annotations, unlike the previous metrics. They directly reflect basic properties of the supervoxel methods, such as the temporal extent of generated supervoxels. 


\subsubsection{Explained Variation (EV)}
\label{subsubsec:EV}

The metric is proposed in \cite{MoPrWaCVPR2008} and it considers the supervoxels 
as a compression method of a video (Sect.~\ref{sec:good}):
\begin{align}
\text{EV}(\mathbf{s}) = \frac{\sum_i (\mu_i - \mu)^2}{\sum_i (x_i - \mu)^2}
\enspace,
\label{eq:EV}
\end{align}
where $x_i$ is the color of the video voxel $i$, $\mu$ is the mean color of all voxels in a video and $\mu_i$ is the mean color of the supervoxel that contains voxel $i$.  \cite{ErSaTeTIP2004} observe a correlation between EV and the human-dependent metrics for a specific object tracking task.

\subsubsection{Mean Size Variation (MSV)}
\label{subsubsec:MSV}

\cite{ChWeIICVPR2013} propose superpixel size variation that measures the size variation of all superpixels in a video (as a set of frames). Here, we extend their metric to measure the size variation of the 2D slices of a supervoxel. MSV is the average score of such variation defined by all supervoxels in a video:
\begin{align}
\text{MSV}(\mathbf{s}) = \frac{1}{n} \sum_{j=1}^n
\sqrt{ 
\frac{\sum_t \left(\left(|s_i^t| - |\hat{s_i}|\right)^2 \mathbbm{1}\left[|s_i^t| > 
0\right]\right)}
{\sum_t \mathbbm{1}[|s_i^t| > 0] - 1} }
\enspace,
\label{eq:MSV}
\end{align}
where $|\hat{s_i}| = \frac{\sum_t |s_i^t|}{\sum_t \mathbbm{1}[|s_i^t| > 0]}$ is 
the average size of 2D slices of a supervoxel. MSV favors the kind of 
supervoxels whose 2D sizes varies minimally over time.

\subsubsection{Temporal Extent (TEX)}
\label{subsubsec:TEX}

TEX measures the average temporal extent of all supervoxels in a video. The measure of supervoxel temporal extent is originally proposed in \cite{XuXiCoECCV2012} as a way to compare different streaming video segmentation methods. Later, \cite{ChWeIICVPR2013} extend the measure by normalizing over the number of frames contained in a video. We also use it here for the evaluation. The metric is defined as follows:
\begin{align}
\text{TEX}(\mathbf{s}) = \frac{1}{nT} \sum_{i=1}^n \sum_{t=1}^T \mathbbm{1}[|s_i^t| > 0]
\enspace.
\label{eq:TEX}
\end{align}

\subsection{Computational Cost}
\label{subsec:cost}

\begin{table}
	\centering
	\includegraphics[width=\linewidth]{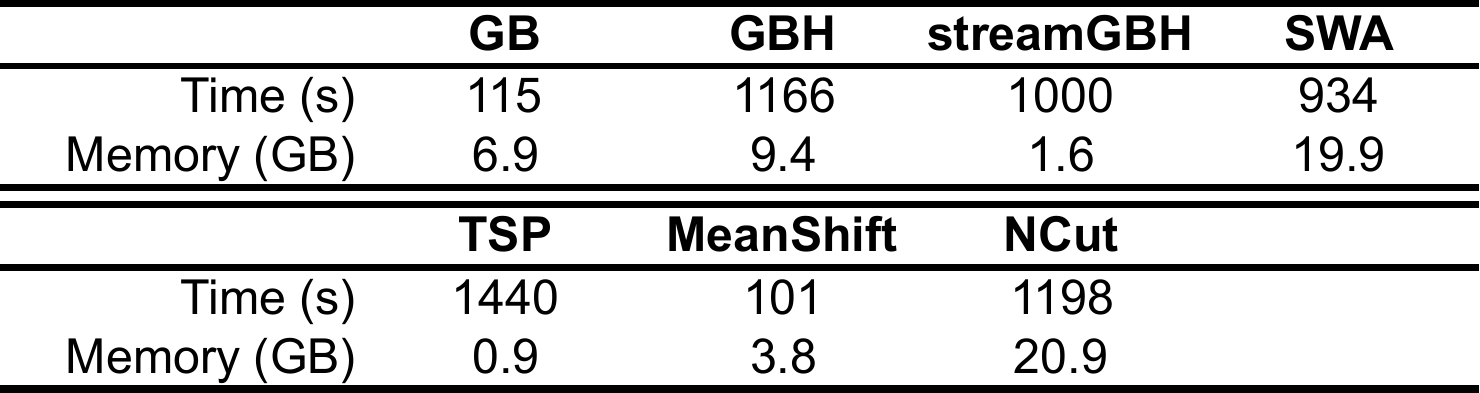}
	\caption{Computational cost.}
	\label{tab:cost}
\end{table}

We report the computational cost of all methods for a typical video with $352 \times 288 \times 85$ voxels---we record the time and peak memory consumption on a laptop featured with Intel Core i7-3740QM @ 2.70GHz and 32GB RAM running Linux, see Table~\ref{tab:cost}. All methods are implemented in C except NCut (Matlab) and TSP (Matlab with MEX). Furthermore, all methods are single threaded except NCut running with 8 threads with resized resolution to $240 \times 160$. 

\subsection{Discussion}
\label{subsec:discussion}

\begin{figure*}
\begin{center}
\includegraphics[width=\linewidth]{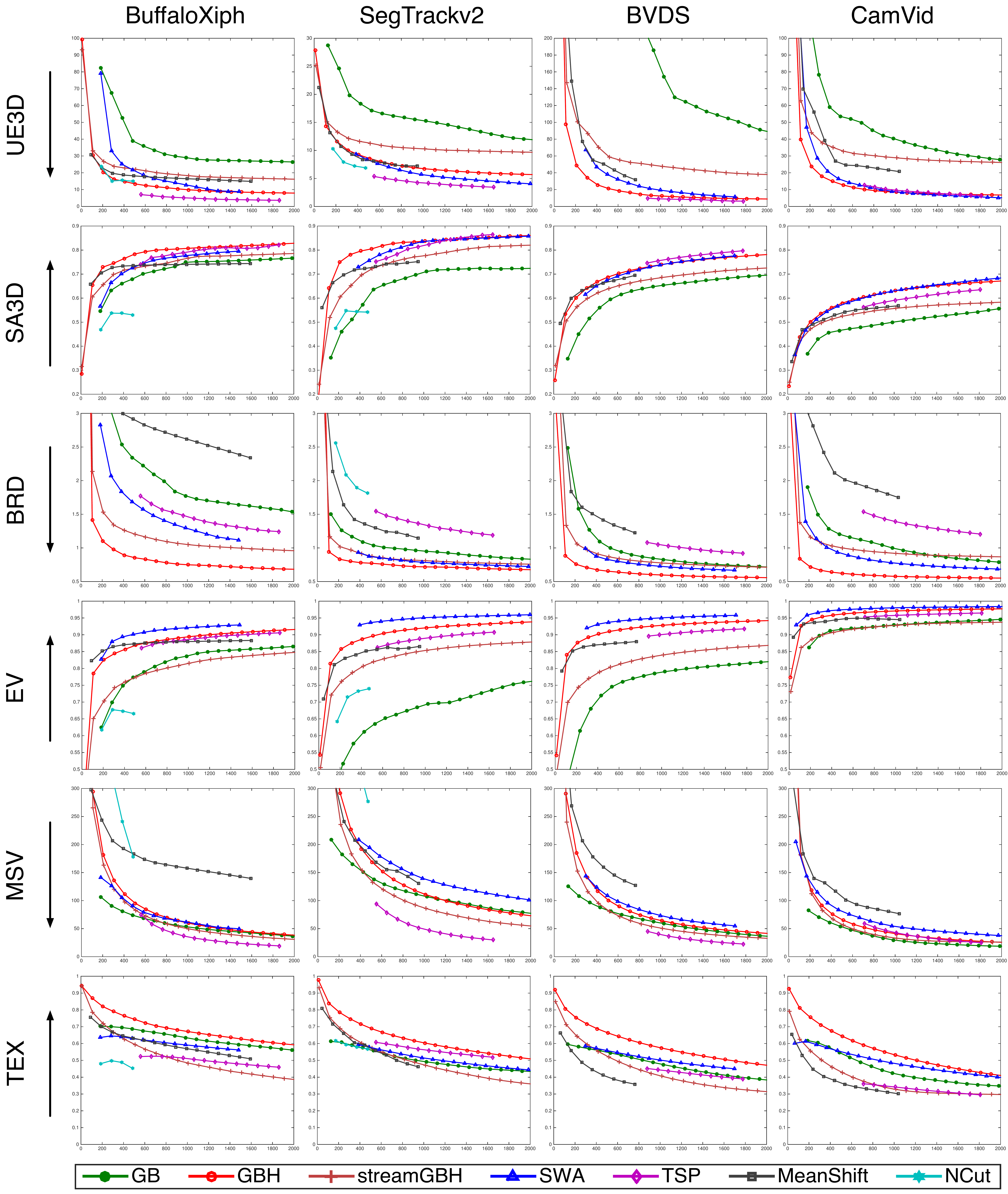}
\caption{Graphs plot the number of supervoxels \textbf{per-video} (x-axis) 
against various metrics (y-axis). Datasets are organized by columns and metrics 
are organized by rows. Black arrows in each row are used to indicate the 
direction of better performance with regard to the metric. Plot ranges along 
the y-axis are aligned for all metrics except UE3D. Plotted dots are the average 
score of linear-interpolated values from all videos in a dataset at the same 
number of supervoxels per-video.}
\label{fig:spv}
\end{center}
\end{figure*}

\begin{figure*}
\begin{center}
\includegraphics[width=\linewidth]{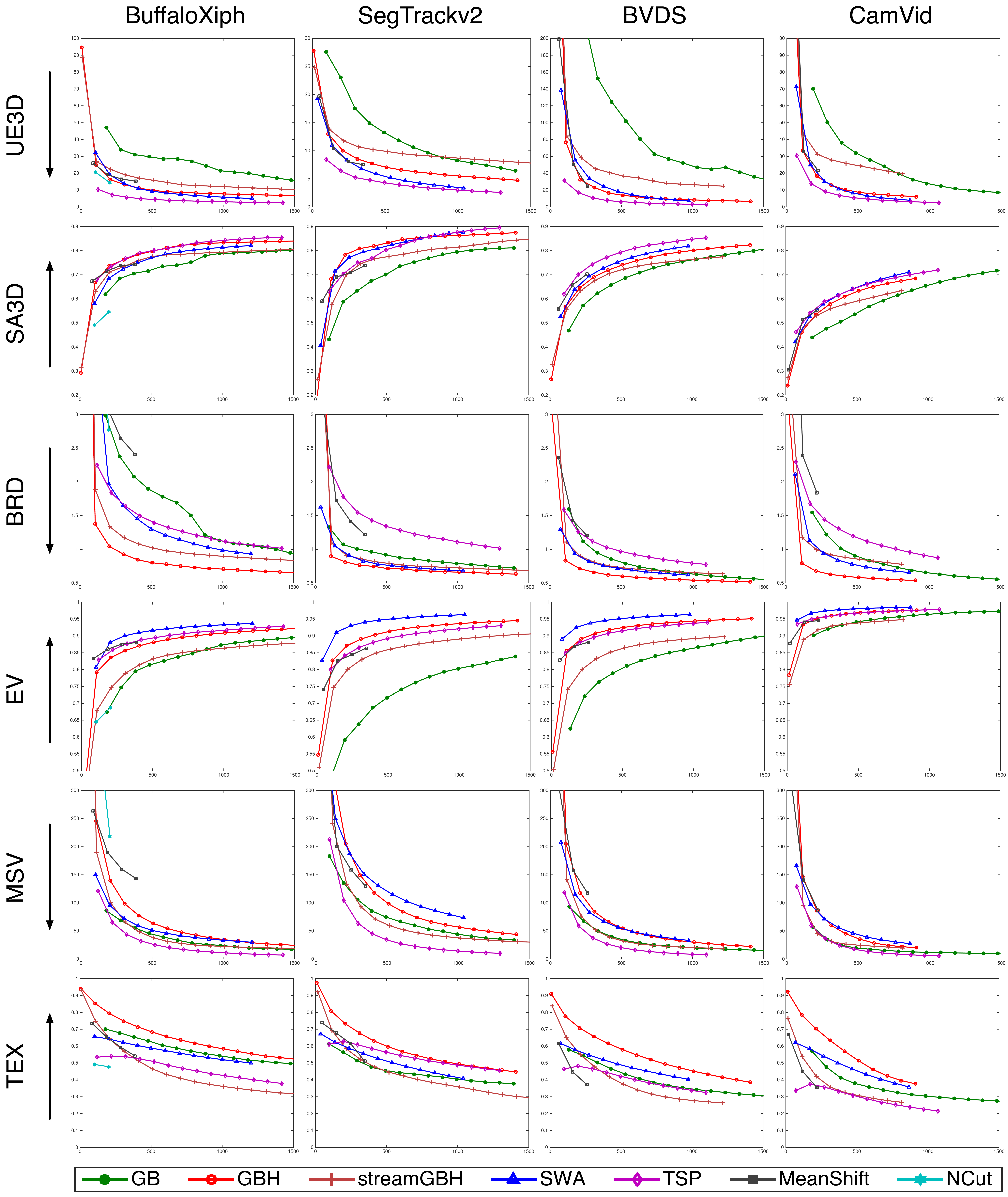}
\caption{Graphs plot the number of supervoxels \textbf{per-frame} (x-axis) 
against various metrics (y-axis). Datasets are organized by columns and metrics 
are organized by rows. Black arrows in each row are used to indicate the 
direction of better performance with regard to the metric. Plot ranges along 
the y-axis are aligned for all metrics except UE3D. Plotted dots are the average 
score of linear-interpolated values from all videos in a dataset at the same 
number of supervoxels per-frame.}
\label{fig:spf}
\end{center}
\end{figure*}

\begin{figure}
\begin{center}
\includegraphics[width=\linewidth]{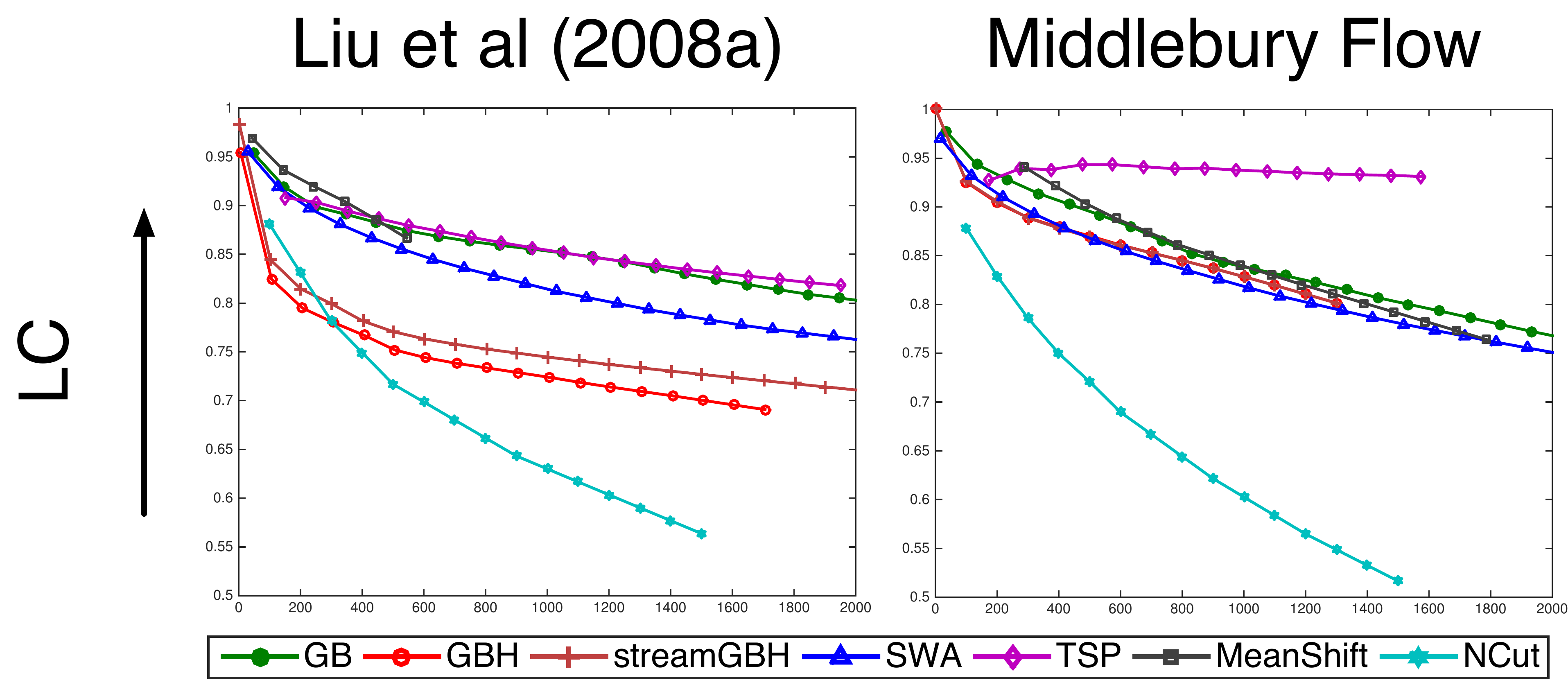}
\caption{Plots for Label Consistency (LC) against the number of supervoxels 
\textbf{per-video} (x-axis). Black arrow indicates the direction of better performance. Plotted dots are the average score of linear-interpolated values from all videos in a dataset at the same number of supervoxels per-video.}
\label{fig:LC_spv}
\end{center}
\end{figure}

\begin{figure}
\begin{center}
\includegraphics[width=\linewidth]{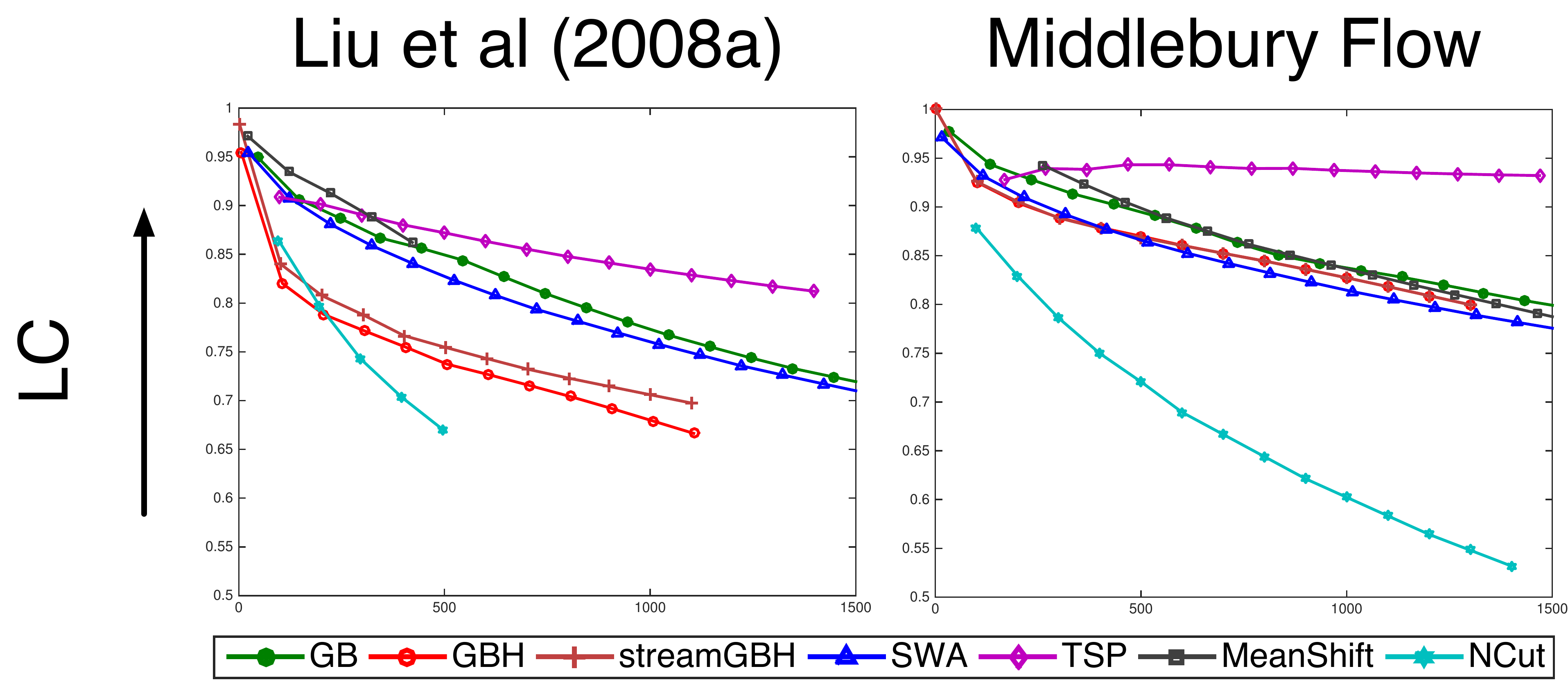}
\caption{Plots for Label Consistency (LC) based on the number of supervoxels 
\textbf{per-frame} (x-axis). Black arrow indicates the direction of better performance. Plotted dots are the average score of linear-interpolated values from all videos in a dataset at the same number of supervoxels per-frame.}
\label{fig:LC_spf}
\end{center}
\end{figure}

\begin{figure*}
\begin{center}
\includegraphics[width=0.9\linewidth]{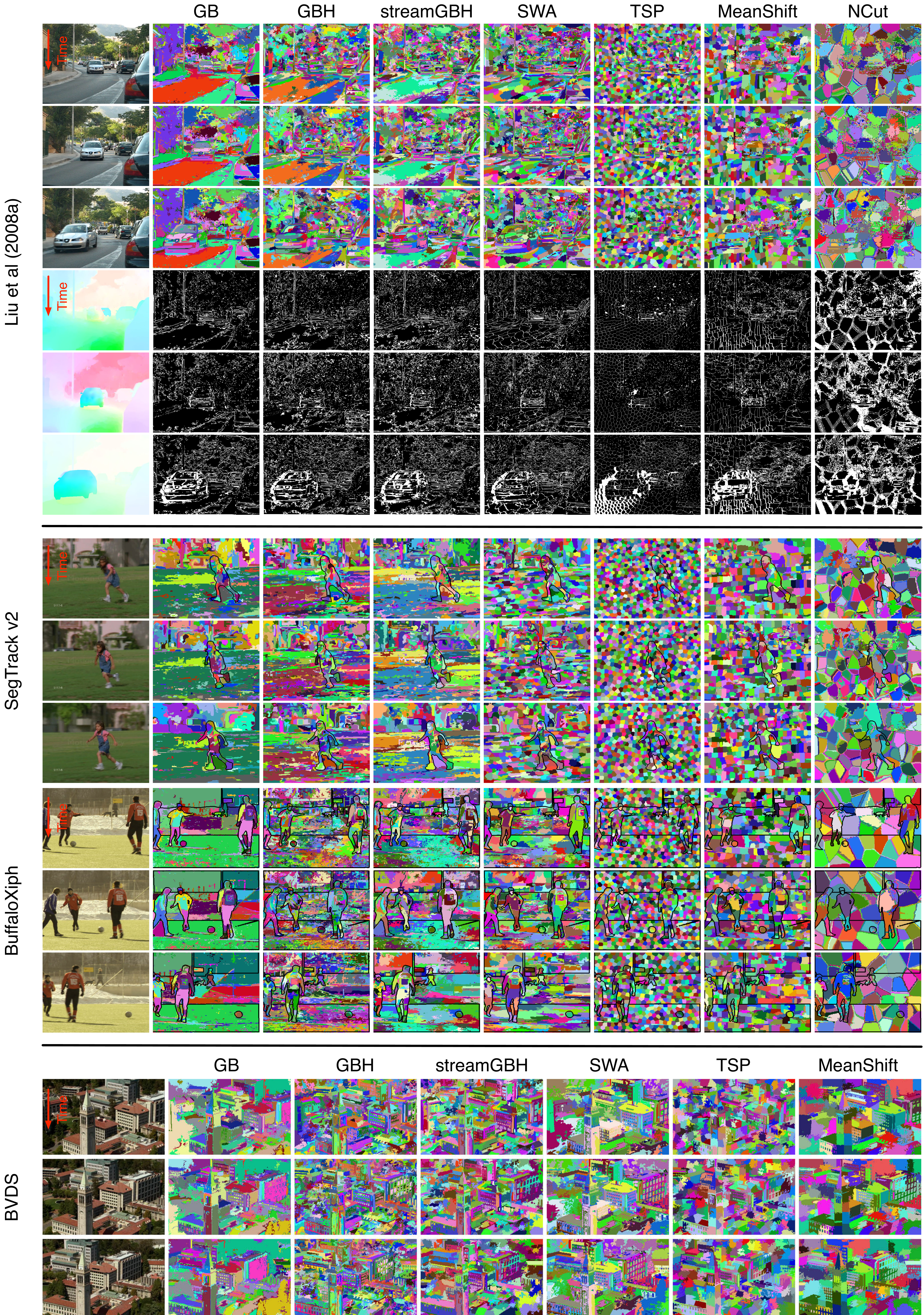}
\caption{Visual comparative results of the seven methods on videos. Each 
supervoxel is rendered with its distinct color and these are maintained over 
time. We recommend viewing these images zoomed on an electronic display. In the 
top part, we show a video from \cite{LiFrAdCVPR2008} where label consistency is 
computed and shown in black and white (white pixels indicate inconsistency with respect to groundtruth flow). In the middle part, we show videos from SegTrack v2 and BuffaloXiph, where groundtruth object boundaries are drawn in black lines. We show a video from BVDS on the bottom.}
\label{fig:montage}
\end{center}
\end{figure*}

We evaluate seven methods over six datasets by the metrics defined above. We focus the evaluation in the range of 0 to 2000 supervoxels per-video (Fig.~\ref{fig:spv} and Fig.~\ref{fig:LC_spv}) as well as 0 to 1500 supervoxels per-frame (Fig.~\ref{fig:spf} and Fig.~\ref{fig:LC_spf}). We do the best to accommodate all methods in the above range, but not all methods can generate the full range of plots (e.g. Mean Shift requires huge memory to generate over 500 supervoxels per-frame for a typical video). The visualization of supervoxel segmentations can be found in Fig.~\ref{fig:montage}. For the rest of this section, we first discuss the choice of two plot bases in Sect.~\ref{subsubsec:PlotBases}, then conclude our findings in Sect.~\ref{subsubsec:Findings}.

\subsubsection{Plot Bases}
\label{subsubsec:PlotBases}

We plot the results with two types of plot bases, namely the number of 
supervoxels per-video and per-frame. We summarize the rationale below, which 
basically distinguishes the two bases according to how space and time are 
treated.  

\noindent \textbf{Number of Supervoxels Per-Video (spv).} In the earlier version 
of our paper \citep{XuCoCVPR2012}, the number of supervoxels per-video is used 
for plotting figures of metric scores.  Here, time is considered as an 
analogous, third dimension and treated accordingly, as in the definition of 
supervoxels.  Hence, for example, one can consider this as a means of evaluating the 
compression-rate of a video as a whole.  However, it may incorrectly relate 
videos of different lengths.

\noindent \textbf{Number of Supervoxels Per-Frame (spf).} \cite{ChWeIICVPR2013} 
use the number of supervoxels per-frame (in this case, it is the same as the 
number of superpixels per-frame) in their evaluation. The mindset behind that 
differentiates the time dimension in a video from the spatial dimensions, such 
that the plot basis is not subject to different video lengths or motion.  
However, this approach fails to account for the temporal qualities of 
supervoxels---a good superpixel method can do well.  For example, UE3D 
degenerates to UE2D for a superpixel method because it has perfect temporal 
boundaries.

\noindent \textbf{Summary.} We hence present plots against both bases and we 
discuss their comparisons.

\subsubsection{Top Performing Methods}
\label{subsubsec:Findings}

The metrics using human annotations, namely UE3D, SA3D, BRD and LC, reflect 
different preferences for supervoxels (see Sect.~\ref{sec:good}). A perfect 
segmentation can have all the best scores with respect to these metrics, while, 
often, a typical segmentation has its strengths in a subset of the metrics 
(recall the example in Fig.~\ref{fig:Box}).  Therefore, we organize our findings 
of the top performing methods by each metric and discuss the differences among 
datasets, if any. Recall that our choices of datasets in 
Sect.~\ref{sec:datasets} represent many different types of video data (e.g.  
SegTrack v2 has only foreground object labels, and BuffaloXiph has pixel-level 
semantic class labels).  Below we list the key results.

\noindent \textbf{UE3D.} For most cases, TSP has the best performance followed 
by SWA and GBH. The three methods have nearly the same good performance on 
CamVid for spv in Fig.~\ref{fig:spv}. However, TSP stands out when evaluating 
for spf in Fig.~\ref{fig:spf}.

\noindent \textbf{SA3D.} GBH performs best on BuffaloXiph for spv, whereas TSP 
performs best when plotted by spf. GBH, SWA and TSP perform almost equally well on 
SegTrack v2. TSP performs best on BVDS, where annotators are instructed to label 
all objects on sampled frames of a video. SWA and GBH perform equally best on 
CamVid for spv, but when plotting by spf, SWA and TSP perform the best.

\noindent \textbf{BRD.} GBH is the clear winner method in this metric, and following that are streamGBH and SWA. GB has a faster trend to approach GBH than streamGBH on CamVid and BVDS for spf. 

\noindent \textbf{LC.} TSP (the only method uses optical flow in the 
implementations we use) has the best performance and there is a clear performance gap on Middlebury Flow, where videos only have two frames (Fig~\ref{fig:LC_spv} and \ref{fig:LC_spf}). Furthermore, unlike the other methods, the performance of TSP dose not dramatically decrease when spv and spf increase on Middlebury Flow. 

\noindent \textbf{EV.} SWA has the overall best performance and followed by GBH and TSP. 
GBH ranks better than TSP on BuffaloXiph for spv, but the ordering swapped when plotting against spf.

\noindent \textbf{MSV.} TSP has the best performance followed by streamGBH and GB except on CamVid for spv, where GB  performs the best. 

\noindent \textbf{TEX.} GBH has the longest temporal extent for both spv and spf within the range we plotted. We note \cite{ChWeIICVPR2013} show that TSP has better performance than GBH in a different spectrum of spf on \cite{LiFrAdCVPR2008} and SegTrack \citep{TsFlReBMVC2010}.

Over all seven methods, GB and Mean Shift are the most efficient in time. Interestingly, neither GB nor Mean Shift performs best in any of the human annotation related quality measures---there is an obvious trade-off between the computational cost of the methods and the quality of their output (in terms of our metrics). 

We have focused on the facts here.  Although understanding \textit{why} these 
various algorithms demonstrate this comparative performance is an ultimate goal 
of our work, it is beyond the scope of this paper and would require a 
substantionally deeper understander of how space and time relate in video 
analysis.  To overcome this limitation, we map these comparative performances 
onto specific problem-oriented needs in the Conclusion (Sect. 
\ref{sec:conclusion}).

\section{Supervoxel Classification}
\label{sec:classification}

\begin{figure}
\begin{center}
\includegraphics[width=\linewidth]{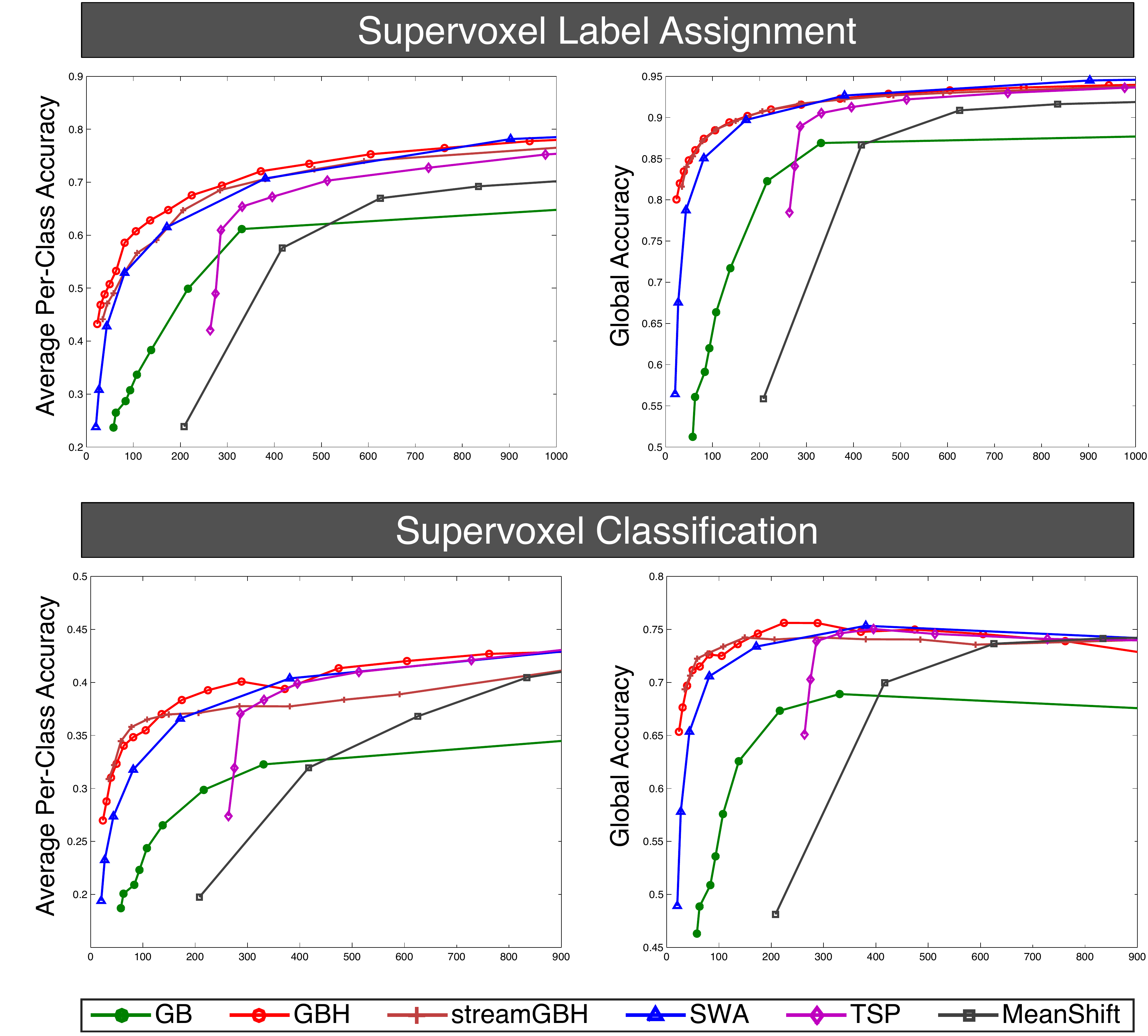}
\caption{Plots on the top are the pixel-level average per-class accuracy (left) and 
global accuracy (right) for both training and testing sets when supervoxels 
directly take groundtruth labels (the most frequent ones in volumes). Plots on 
the bottom are the pixel-level classification performance on the test set with SVMs trained on supervoxels. We show the plots in the range of 100 to 900 supervoxels every 100 frames (x-axis). The plotted dots are from actual segmentations rather than interpolated values. We note that \cite{BrShFaECCV2008} report 53.0\% average per-class and 69.1\% global accuracy using random forests trained on pixels with both appearance and geometric cues, where we only use appearance cues with supervoxels.}
\label{fig:Classification}
\end{center}
\end{figure}

\begin{figure*}
\begin{center}
\includegraphics[width=\linewidth]{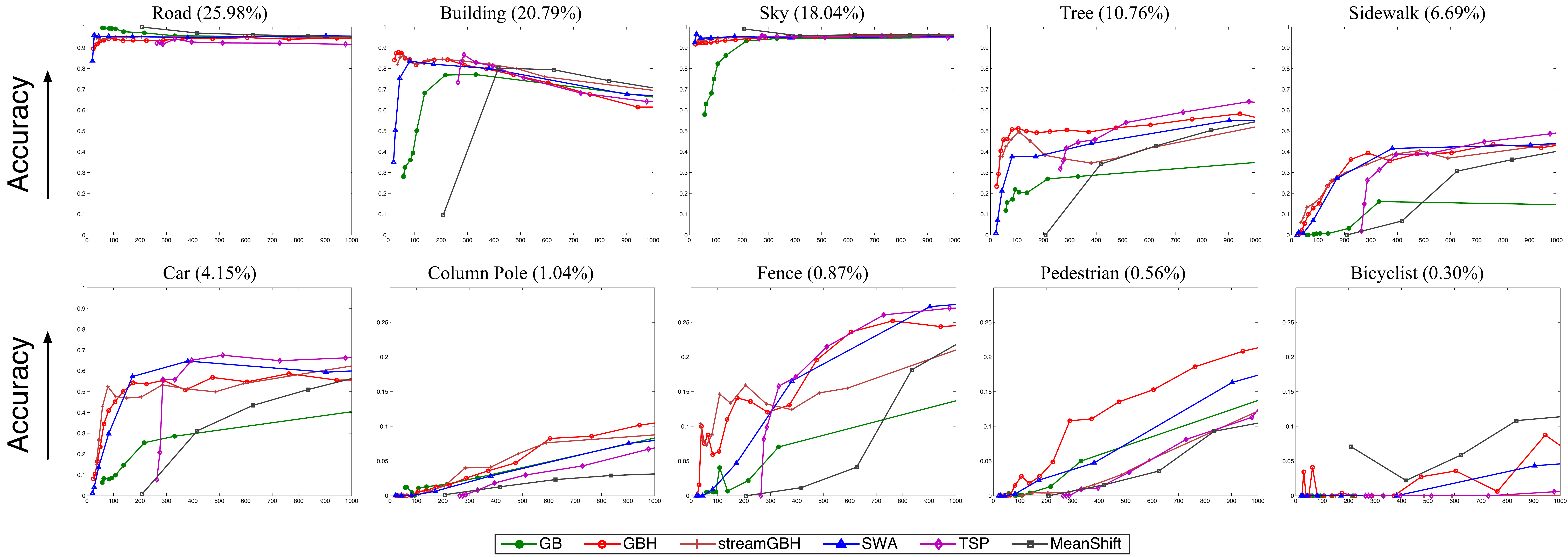}
\caption{Pixel-level labeling accuracy for each semantic class in the CamVid 
dataset, where the percentages of total pixels for each class are shown on top. 
All plots are shown in the range of 0 to 1000 supervoxels every 100 frames 
(x-axis). The first six plots (horizontal) are plotted with an accuracy range 
from 0 to 1, and the other plots are from 0 to 0.3. We do not show the class \textit{Sign Symbol (0.17\%)} here due to its low accuracy for all methods.}
\label{fig:perClass}
\end{center}
\end{figure*}

In this section, we evaluate the supervoxel methods in a particular application:  supervoxel semantic label classification. We use this application as a proxy to various high-level video analysis problems. For example, superpixel classification scores are frequently used as the unary term when building subsequent graphical model for scene understanding in images, e.g., \cite{GoFuKoICCV2009}. We use the CamVid dataset for this task due to its widely use in semantic pixel labeling in videos. Recall that CamVid has videos over ten minutes and labeled frames with 11 semantic classes at 1Hz, such as building, tree, car and road. We follow the standard training/test split: two daytime and one dusk sequence for training, and one daytime and one dusk sequence for testing. We process the videos into supervoxel segmentations as described in Sect.~\ref{subsec:processing}. We use all supervoxel methods except for the NCut method because of its high memory requirement for these CamVid data, which rendered the size of the supervoxels too large to train meaningful classifiers.

\noindent \textbf{Supervoxel Features.} \cite{TiLaIJCV2010} extract a set of low-level features on superpixels and use the supervoxels generated by \cite{GrKwHaCVPR2010} for their video parsing on CamVid dataset. We apply a similar set of features with some modifications to suit for our task. 
We first dilate the 2D slices of supervoxels by 3 pixels and then extract features histograms from supervoxel volumes. To be specific, we compute histogram of textons\footnote{\url{http://www.robots.ox.ac.uk/~vgg/research/texclass/filters.html}} and dense SIFT descriptors with 100 dimensions each. We also compute two types of color histograms, RGB and HSV, with 8 bins each channel. We describe the location of a supervoxel volume by averaging the distances of bounding boxes of its 2D slices to image boundaries. In addition to image features, we calculate dense optical flow and quantize flows in a supervoxel volume to 8 bins each according to vertical and horizontal velocity, and speed magnitude. Note that the way we extract the feature histograms is different than \cite{TiLaIJCV2010}, where they use one representative superpixel of a supervoxel (the 2D slice with largest region). We think that the volume has better potential to represent the change of a supervoxel over time. We also note that more sophisticated video features can be added to supervoxel volumes such as dense trajectories \citep{WaKlScIJCV2013} and HOG3D \citep{KlMaScBMVC2008}. However, for a fair comparison of supervoxel methods, we stick to the dense image features and optical flow in order to prevent favoring one supervoxel method than another.

\noindent \textbf{Supervoxel Labels.} We assign a supervoxel with the most 
frequent groundtruth label occur in its volume and ignore supervoxels that fail 
to touch groundtruth frames (labeled at 1Hz on CamVid). We note that this step 
is distinct from most image superpixel classification work, e.g., 
\cite{GoFuKoICCV2009}, since videos are often sparsely labeled while images are 
densely labeled. Therefore, this step may introduce more noise in both training 
and testing than the image superpixel classification work, and it is closely 
related to two of our benchmark metrics---UE3D in Sect.~\ref{subsec:UE3D} and 
SA3D in Sect.~\ref{subsec:SA3D}. We apply the pixel-level average per-class 
accuracy and global pixel accuracy to evaluate this \textit{supervoxel label 
assignment} step and the top part in Fig.~\ref{fig:Classification} shows the performance for all six methods in the experiment. Rather than using linear interpolated values as in Fig.~\ref{fig:spv} to \ref{fig:LC_spf}, the plotted dots here map to actual segmentations generated by a single run of the algorithm over the dataset, and the plot basis is the number of supervoxels for every 100 frames. 

\noindent \textbf{Classification Performance.} Finally, we use linear SVMs\footnote{\url{http://www.csie.ntu.edu.tw/~cjlin/libsvm/}} on supervoxels to get the classification results on the test set. The output segmentations are for the entire video but we evaluate only on the labeled frames. We again show the performance in terms of pixel-level average per-class and global accuracy in the bottom part in Fig. ~\ref{fig:Classification} with the number of supervoxels ranging from less that 100 to more than 900 every 100 frames. To compare with pixel-based image segmentation method, we note that \cite{BrShFaECCV2008} report 53.0\% average per-class and 69.1\% global accuracy by using both appearance and geometric cues. The supervoxel-based methods with our setup in general achieve a better global pixel performance but a worse average per-class accuracy (e.g. 500 supervoxels in Fig. \ref{fig:Classification}) with respect to the range of supervoxel numbers we sampled for the evaluation. We suspect that some classes with small regions, such as \textit{sign symbol} and \textit{bicyclist}, become too small to capture when we scale the videos down to a much lower resolution (a quarter of the original) to accommodate all six supervoxel methods. 

Fig.~\ref{fig:perClass} shows the pixel-level labeling accuracy for each class 
in the dataset. For large classes, such as \textit{road} and \textit{sky}, all 
methods perform almost equally well regardless of the change of supervoxel 
numbers, except the performance for \textit{building}, which rises then falls.  
This rise and fall results in a decrease in the overall global performance (see 
bottom right in Fig.~\ref{fig:Classification}).  We explain this rise and fall 
behavior of the building class due to the overall scale-varying texture of 
buildings and the challenge to learn classifiers on them that perform equally 
well at different scales; for example, smaller supervoxels will cover small 
portions of buildings, say windows or bricks, which have distinct visual 
characteristics, yet a single classifier is to be learned (in our evaluation).
For other classes the performance increases when adding more supervoxels, and 
different methods have distinct performance on different classes. For example, 
GBH leads the score on \textit{pedestrian} while TSP and SWA are the methods of 
choice on \textit{car}.  Further investigation is needed to better understand 
these nuances.

\begin{figure*}
\begin{center}
\includegraphics[width=0.9\linewidth]{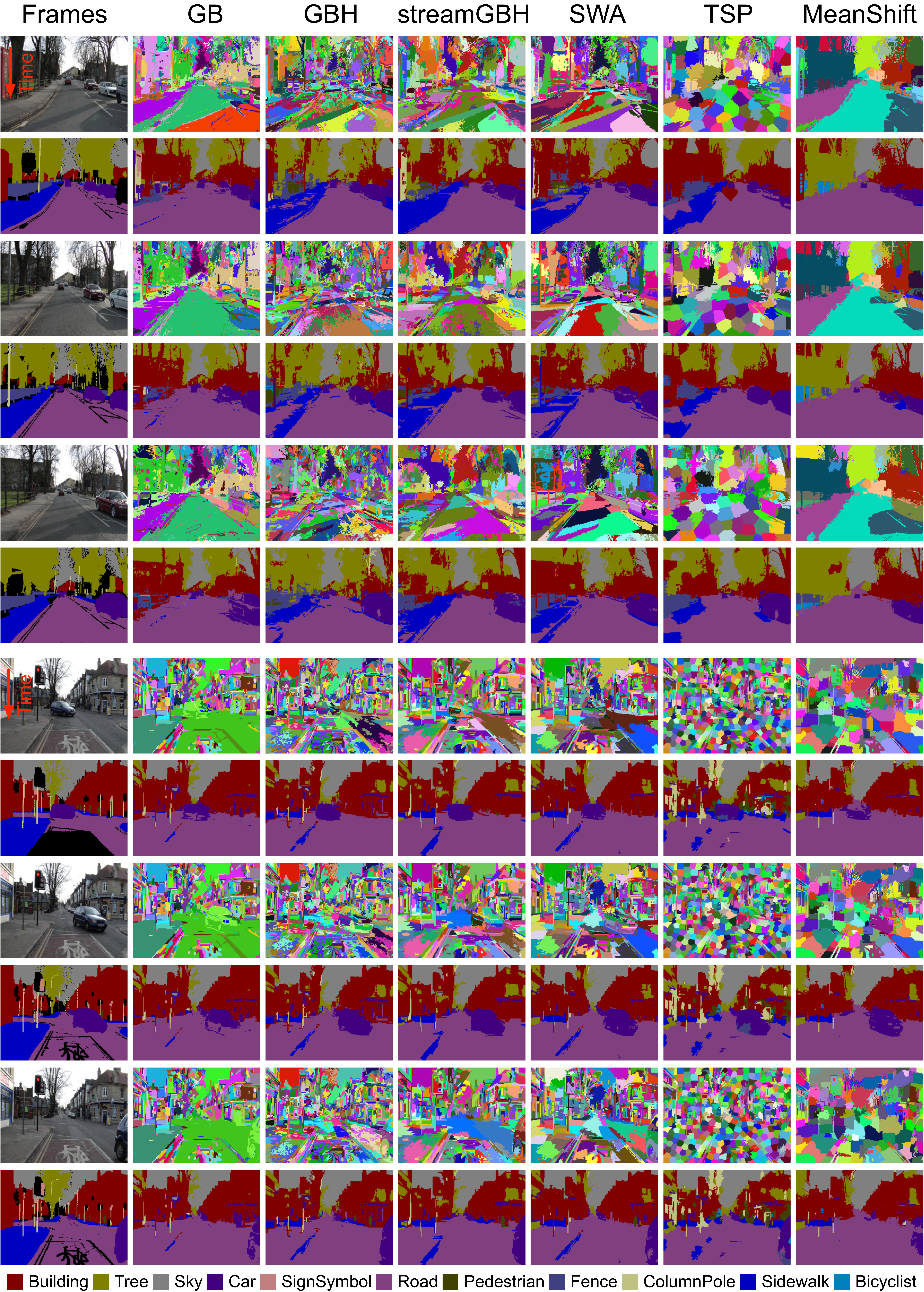}
\caption{Example results on two short clips from the CamVid daytime test video. 
Images in the first column are video frames and groundtruth labels and the 
remaining columns are individual methods with supervoxel segmentation and semantic labeling on supervoxels.}
\label{fig:seq4}
\end{center}
\end{figure*}

\begin{figure*}
\begin{center}
\includegraphics[width=0.9\linewidth]{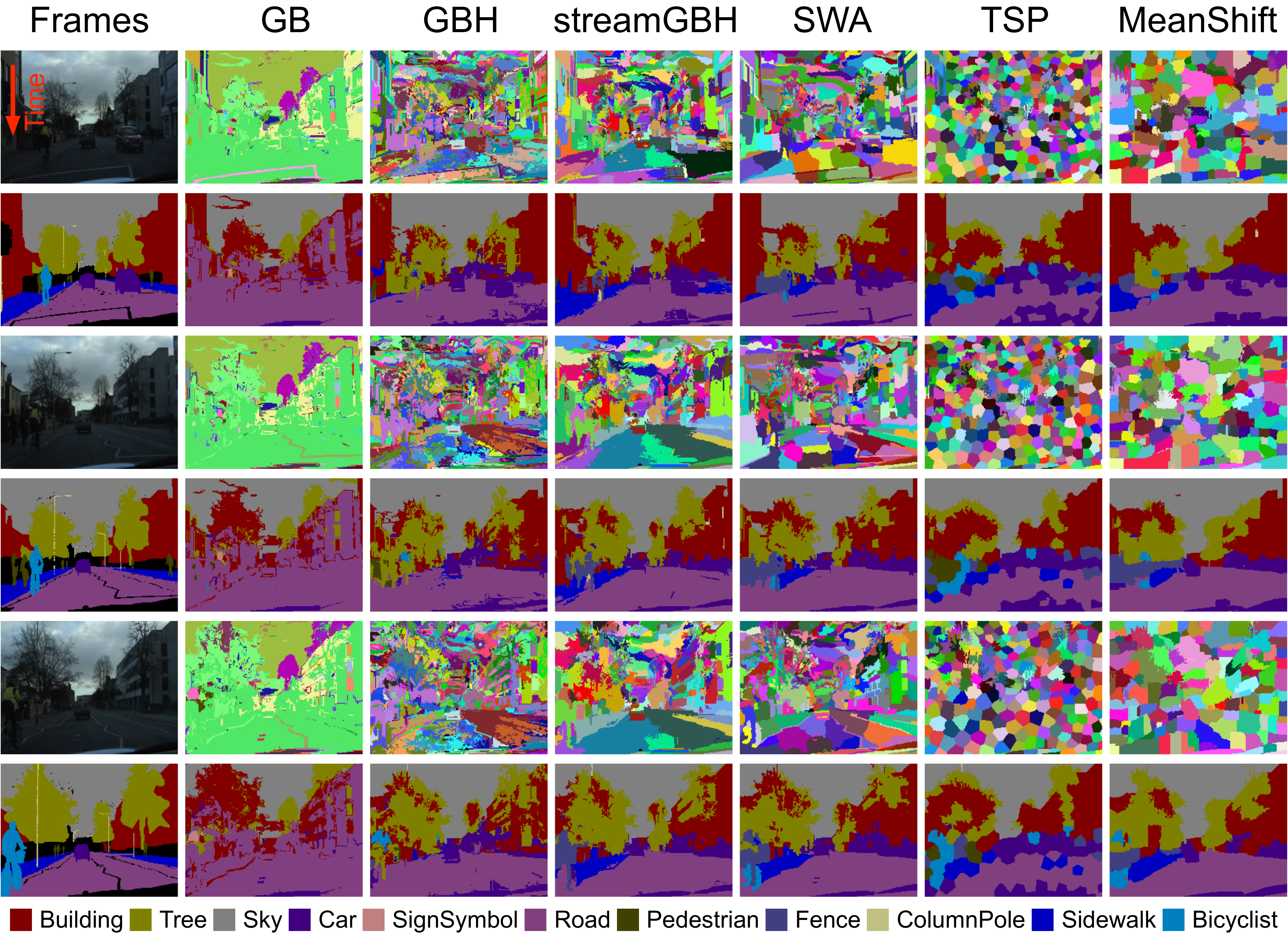}
\caption{Example results on a clip from the CamVid dusk test video. Images in 
the first column are video frames and groundtruth labels and the remaining columns are individual methods with supervoxel segmentation and semantic labeling on supervoxels.}
\label{fig:seq5}
\end{center}
\end{figure*}

Fig.~\ref{fig:seq4} shows visual comparison of six methods on two clips from the daytime test video. Although GB and Mean Shift successfully segment the \textit{sidewalk} in the supervoxel segmentation, they miss a large portion of the \textit{sidewalk} in the labeling, while the other methods capture it well. The \textit{tree} tends to be better labeled by GBH. All methods segment the moving cars well. However none of the method get the small \textit{sign symbol} in the second clip. We also show the results during dusk in Fig.~\ref{fig:seq5}. GB works poorly here; the greedy algorithm of GB is highly sensitive to local color thus it easily produces large incorrect segments. TSP visually segments \textit{bicyclist} well regardless the incorrect boundaries. We think this is due to the compact shape of supervoxels that TSP generated can better track the superpixels on the bicyclist and prevent easily merging with other large segments such as \textit{sidewalk}, \textit{tree} and \textit{road}. However, it also brings more fragmented segments on large smooth regions, such as \textit{road} and \textit{sidewalk} and weak boundary accuracy.

Overall, GBH, SWA and TSP achieve equally strong performance in the supervoxel 
classification experiment (see Fig. \ref{fig:Classification}), and, again, they 
are the top performing ones in terms of our benchmark evaluation in Sect. 
\ref{sec:benchmark}. Methods such as GB and MeanShift have poor classification 
performance also perform less well on the benchmark metrics. For the streaming 
methods, streamGBH achieves very similar performance to its full-video 
counterpart GBH.

\section{Conclusion}
\label{sec:conclusion}

We have presented a thorough evaluation of seven supervoxel methods including 
both off-line and streaming methods on a set of seven benchmark metrics designed 
to evaluate supervoxel desiderata as well as the recognition performance on a 
particular application. Samples from the datasets segmented under all seven 
methods are shown in Fig. \ref{fig:montage}, Fig. \ref{fig:seq4}, and Fig.  
\ref{fig:seq5}. These visual results convey the overall findings we have 
observed in the quantitative experiments. GBH, SWA and TSP are the 
top-performers among the seven methods in both our benchmark evaluation and the 
classification task. They all share a common feature in that they perform well 
in terms of segmentation accuracy, but they comparatively vary in performance in 
regard to the other metrics. 
GBH captures object boundaries best making it well suited for video analysis 
tasks when accurate boundaries are needed, such as robot manipulation.
SWA has the best performance in the explained variation metric, which makes it 
most well-suited for compression applications.
TSP follows object parts and achieves the best undersegmentation error making it 
well-suited for fine-grained activity analysis and other high-level video 
understanding problems. 
It seems evident that the main distinction behind the best offline methods, 
namely GBH and SWA, is the way in which they both compute the hierarchical 
segmentation.  Although the details differ, the common feature among the two 
methods is that during the hierarchical computation, coarse-level aggregate 
features replace or modulate fine-level individual features. In contrast, TSP 
processes a video in a streaming fashion and also produces supervoxels that are 
the most compact and regular in shape.  These differences suggest a 
complementarity that has the potential to be combined into a new method, which 
are currently investigating.

In this paper, we have explicitly studied the general supervoxel desiderata regarding a set of proposed benchmark metrics including both human annotation dependent and independent ones. In addition, we compare the supervoxel methods in a particular application---supervoxel classification that evaluates methods in a recognition task, which we consider to be a proxy to various high-level video analysis tasks in which supervoxels could be used. A strong correlation is presented between the benchmark evaluation and the recognition task. Methods, such as GBH, SWA and TSP, that achieve the top performance in the benchmark evaluation also perform best in the recognition task. The obvious question to ask is how well will the findings translate to other application-specific ones, such as tracking and activity recognition. A related additional point that needs further exploration for supervoxel methods is the modeling of the relationship between spatial and temporal domains in a video. We plan to study these important questions in future work.

\begin{acknowledgements}
This work was partially supported by the National Science Foundation CAREER grant (IIS-0845282), the Army Research Office (W911NF-11-1-0090) and the DARPA Mind's Eye program (W911NF-10-2-0062). We are grateful to the authors of the code and datasets that we have relied upon in this study, and we are grateful to the reviewers' comments, which have greatly improved this paper.
\end{acknowledgements}

\bibliographystyle{spbasic}      
\bibliography{LibsvxRev}   

\end{document}